\documentclass[conference]{IEEEtran}

\IEEEoverridecommandlockouts   
\usepackage{cite}
\usepackage{amsmath,amssymb,amsfonts}
\usepackage{graphicx}
\usepackage{textcomp}
\usepackage{xcolor}
\usepackage{bm}
\usepackage{mathtools}
\usepackage{multirow}
\usepackage{array}
\usepackage{booktabs}
\usepackage{url}
\usepackage{enumitem}

\usepackage{bbold}
\usepackage{dsfont}
\usepackage{isomath}
\usepackage{arydshln}
\usepackage{stmaryrd}
\usepackage{pbox}
\usepackage{colortbl}
\usepackage{tikz}
\usepackage{pgf}
\usepackage[normalem]{ulem}

\usepackage[caption=false,font=footnotesize]{subfig}

\newcommand{\stkout}[1]{{\ifmmode\text{\sout{\ensuremath{#1}}}\else\sout{#1}\fi}}

\usepackage[hidelinks]{hyperref}

\title{CountFormer: A Transformer Framework for Learning Visual Repetition and Structure in Class-Agnostic Object Counting}

\author{
\IEEEauthorblockN{
Md Tanvir Hossain\textsuperscript{1}, 
Akif Islam\textsuperscript{2}, 
Mohd Ruhul Ameen\textsuperscript{3}
}
\IEEEauthorblockA{
\textsuperscript{1,2,3}Department of Computer Science and Engineering, University of Rajshahi\\
Rajshahi, Bangladesh\\
\textsuperscript{1}mth\_cse@ru.ac.bd, 
\textsuperscript{2}iamakifislam@gmail.com, 
\textsuperscript{3}ameensunny242@ru.ac.bd
}
}

\begin{document}
\maketitle

\vspace{0.5cm}
\begin{center}
\small Accepted at the 2026 IEEE 2nd International Conference on Quantum Photonics, Artificial Intelligence and Networking (QPAIN 2026). Copyright IEEE.
\end{center}
\vspace{0.5cm}

\begin{abstract}
Humans can often count unfamiliar objects by observing visual repetition and how parts compose a whole, rather than relying solely on object categories. On the other hand, many counting models that are not given example object patches at test time (i.e., exemplar-free models) struggle in such situations and may overcount when objects contain symmetric components, repeated substructures, or partial occlusion. We introduce \textbf{CountFormer}, a controlled adaptation of a standard density-regression framework inspired by CounTR, in which we replace the image encoder with a self-supervised vision foundation model (\textbf{DINOv2}). The resulting transformer features are augmented with explicit two-dimensional positional embeddings and decoded using a lightweight convolutional network to generate a density map whose integral gives the final object count. Rather than proposing a new counting architecture or objective, our goal is to study whether foundation-based representations influence structural consistency under a strictly exemplar-free inference setting. On FSC-147 (Few-Shot Counting 147), CountFormer achieves competitive performance among existing exemplar-free methods under the official benchmark protocol (MAE 19.06, RMSE 118.45 on the test set). Qualitative analysis indicates fewer part-level overcounting errors in certain structurally complex objects (e.g., glasses), while aggregate MAE and RMSE remain broadly consistent with prior approaches. A diagnostic sensitivity analysis further reveals that evaluation metrics are strongly influenced by a small number of extreme high-density scenes. When four such dense cases are excluded for analysis purposes only, test MAE decreases to 13.14 and RMSE to 33.05, highlighting how a small number of extreme high-density scenes disproportionately contribute to the squared error. Overall, the results highlight the role of representation quality in shaping structural consistency in exemplar-free object counting.
\end{abstract}


\section{Introduction}

We often pause to admire how effortlessly we count things around us as our eyes pick up repeating shapes and patterns—even if we have no idea what class they belong to. Yet when we hand this task over to a machine, the magic fades. Modern computer‐vision systems shine when they are trained to count a specific class of objects: people \cite{Zhang_2016_CVPR}, vehicles \cite{mundhenk2016large}, or other familiar categories. But when they face objects of a completely new type, without any exemplar or prompt, they struggle. Imagine a model staring at a pair of sunglasses and, without a sense of structure, counting each lens as two separate objects. This kind of error reveals a deeper issue: the machine can see shapes, but it cannot always grasp how the parts come together.

Earlier work tried to remedy this by matching patches within an image and counting through similarity \cite{lu2019class}. More recent efforts built on transformers to scan entire scenes and generalize to unseen classes \cite{countr}. Some vision-language systems even substitute exemplars with text prompts \cite{jiang2023clip}. But these approaches often focus on what the object is rather than how its parts are arranged, and they falter when structural coherence matters.

In this work, our design is architecturally close to CounTR, and our goal is not to propose a new counting formulation. Instead, we investigate whether self-supervised foundation features can improve structural robustness in exemplar-free counting, particularly for composite objects where part-level overcounting is common (e.g., glasses). We position \textbf{CountFormer} as a controlled integration of a foundation encoder (DINOv2) and explicit positional fusion within an established density-regression framework. The extracted transformer token features are fused with explicit two-dimensional positional embeddings prior to decoding, and a lightweight convolutional upsampling network produces a continuous density map whose integral yields the final object count.

Our contributions are summarized as follows:
\begin{enumerate}
    \item We present a controlled integration of a DINOv2 vision transformer into a standard exemplar-free density regression framework, architecturally aligned with CounTR, enabling a focused study of how self-supervised foundation features influence structural robustness without altering the loss formulation or inference protocol.
    \item We incorporate a simple two-dimensional positional embedding fusion step prior to density decoding, providing explicit spatial grounding to transformer token representations while keeping the overall architecture lightweight and computationally comparable to prior approaches.
    \item Beyond reporting standard MAE and RMSE on FSC-147, we provide qualitative analyses highlighting reduced part-level overcounting in structurally complex objects and include a diagnostic sensitivity analysis on extreme dense-scene cases.
    \item We explicitly characterize observed failure modes in highly dense scenes with weak inter-object boundaries and clarify evaluation protocol sensitivity without modifying the official benchmark results.
\end{enumerate}

\section{Related Work}

\begin{figure*}[!ht]
    \centering
    \includegraphics[width=\textwidth]{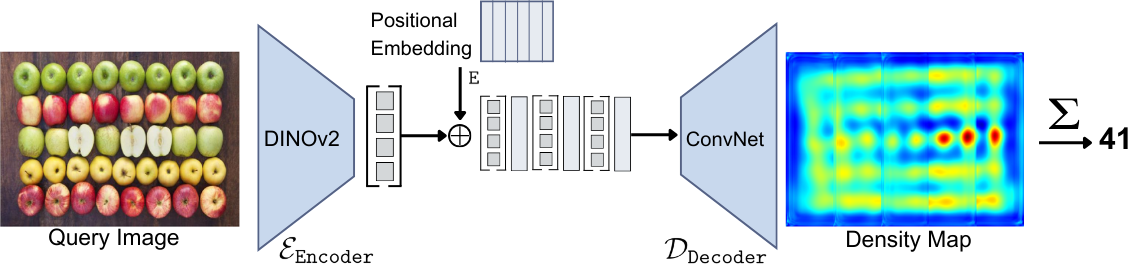}
    \caption{Detailed overview of the proposed model architecture: Our approach utilizes a self-supervised visual encoder, specifically DINOv2, to process the query image. The extracted image features are then summed with the positional embeddings which serve as query vectors. Subsequently, we use a CNN-based decoder to decode these vectors and upscale them to generate the corresponding density map. Finally, the total object count is determined by integrating the values across the density map.}
    \label{fig:architecture}
\end{figure*}

\subsection{Class-Specific Object Counting}
Early works in object counting mainly targeted \textit{class-specific} scenarios, such as crowd counting \cite{Zhang_2016_CVPR} or vehicle counting \cite{mundhenk2016large}.  
These approaches generally fall into two main categories: \textit{detection-based} methods, which identify individual instances through bounding boxes, and \textit{regression-based} methods, which estimate object density maps whose summed pixel values yield the total count \cite{xu2023zero}.  
While highly effective for their target domains, these approaches lack generalization and fail when applied to unseen object types.  
In contrast, humans can effortlessly count diverse objects regardless of category—highlighting a gap between task-specific models and human-like perceptual flexibility.

\subsection{Class-Agnostic and Exemplar-Free Counting}
To overcome this limitation, \textit{class-agnostic} counting methods have emerged, aiming to generalize across categories without retraining.  
The General Matching Network (GMN) \cite{lu2019class} and Few-shot Adaptation and Matching Network (FamNet) \cite{ranjan2021learning} formulated counting as a visual matching problem—estimating similarity between patches within an image or between query and exemplar regions.  
However, these approaches still depend on exemplar patches provided by humans, which limits their scalability and automation.

More recent research has moved toward \textit{exemplar-free} counting.  
RepRPN \cite{ranjan2022exemplar} removed the need for exemplars but primarily detected the dominant object class in an image.  
Vision-language models (VLMs) such as CLIP-Count \cite{jiang2023clip} and CounTX \cite{countx} replaced exemplars with textual prompts, leveraging CLIP \cite{CLIP} to align image and text features through cross-attention.  
While these methods reduced manual supervision, they remained dependent on well-crafted prompts, which introduces linguistic bias and limits fully automatic operation.

\subsection{Representation Learning in Counting}
Beyond architecture design, representation learning plays a crucial role in counting performance.  
CLIP-based methods offer rich semantic understanding but often overlook spatial structure and quantitative relationships—essential for accurate counting.  
Paiss et al. \cite{paiss2022no, paiss2023teaching} demonstrated that CLIP tends to emphasize object identity words in captions while ignoring numerical cues, leading to systematic over- or undercounting.  
This limitation makes CLIP-based models prone to errors in structurally complex objects, such as mistaking both lenses of a pair of glasses as two separate instances.

Attention-based and self-supervised encoders have shown promise in mitigating these issues.  
CounTR \cite{countr} introduced a transformer-based framework that leveraged patch-level attention and positional encoding to model global context, achieving competitive zero-shot results.  
RCC \cite{hobley2022learning} advanced this direction by incorporating a self-supervised DINO-based teacher–student framework for density regression.  
While effective, these encoders still struggled to capture \textit{structural coherence}—the spatial relationships binding multiple parts into a single object.

\begin{figure}[!t]
\centering
\subfloat[Input image (GT = 149 objects)]{
    \includegraphics[width=0.35\columnwidth]{figures/first_page_fig/full_722.png}
}
\hfill
\subfloat[Predicted density map (Count = 149)]{
    \includegraphics[width=0.43\columnwidth]{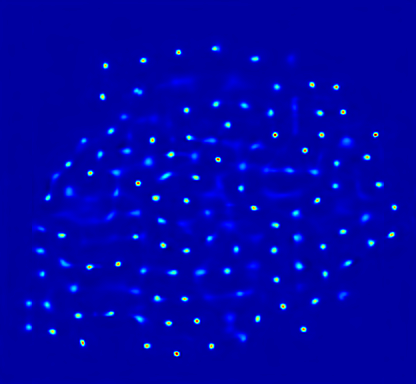}
}
\caption{For the given image (left) with a ground truth of 149 objects, our model produced a density map (right) that was summed to predict an object count of 149.}
\label{fig:lego}
\end{figure}

\subsection{DINOv2 and Structural Awareness}
The recent rise of foundation models such as DINOv2 \cite{dinov2} has transformed self-supervised visual learning.  
Unlike CLIP, which learns from paired image–text data, DINOv2 learns directly from visual information, enabling it to encode both semantic meaning and spatial structure without manual labels.  
This allows DINOv2 to produce feature embeddings that are \textit{structurally aware}—capturing fine-grained relationships between object parts and their spatial organization.  
Such representations are particularly valuable for counting tasks, where recognizing the boundaries and repetition patterns of objects is essential.  

In this work, we leverage DINOv2 as a self-supervised image encoder and fuse its representations with positional embeddings to maintain geometric consistency.  
By doing so, we bridge the gap between semantic abstraction and spatial precision, enabling more reliable and interpretable zero-shot object counting—especially for complex or composite objects.

\begin{figure}[!t]
\centering
\subfloat[Test: Lego]{
    \includegraphics[width=0.45\columnwidth,height=2cm]{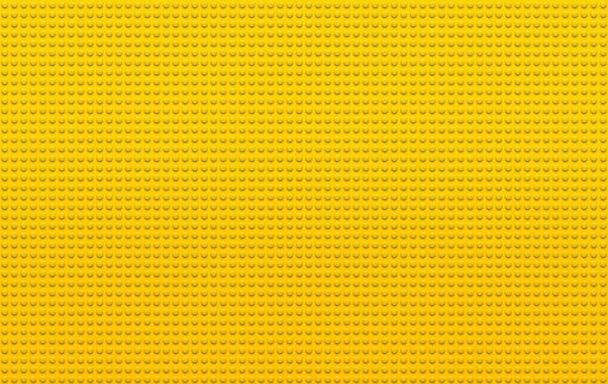}
}\hfill
\subfloat[Test: Pens]{
    \includegraphics[width=0.45\columnwidth,height=2cm]{figures/limitations/pens.jpg}
}\\[4pt]
\subfloat[Validation: Birds]{
    \includegraphics[width=0.45\columnwidth,height=2cm]{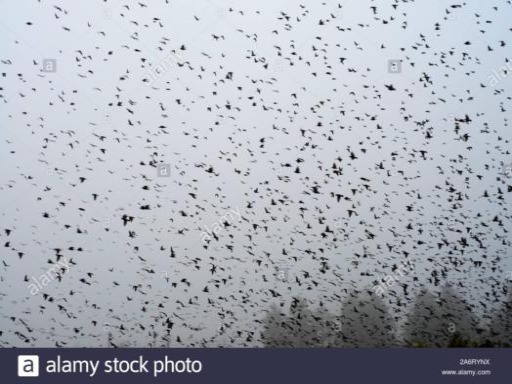}
}\hfill
\subfloat[Validation: Shirts]{
    \includegraphics[width=0.45\columnwidth,height=2cm]{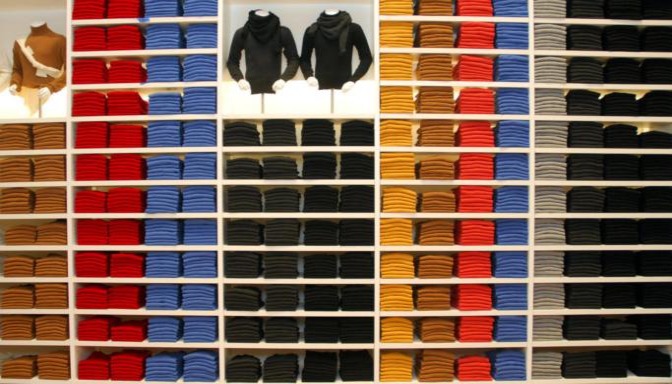}
}
\caption{Failure examples illustrating limitations of our model on images with large object counts and weak inter-object separation. Top row: test set; bottom row: validation set.}
\label{fig:limitations}
\end{figure}

\subsection{Research Gap and Motivation}
Despite steady progress, current class-agnostic counting methods face a persistent limitation: they lack structural awareness.  
Models such as CounTR \cite{countr} and RCC \cite{hobley2022learning} capture global attention or semantic similarity but fail to encode how object parts relate spatially within the image.  
As a result, they often misinterpret composite or repetitive structures—counting object fragments rather than whole instances.  
Existing vision-language models like CLIP \cite{CLIP} further amplify this issue by prioritizing semantics over geometry.  
To bridge this gap, our work introduces a structure-aware zero-shot counting framework built upon DINOv2 \cite{dinov2}, a self-supervised foundation model that inherently preserves spatial coherence.  
By combining DINOv2’s rich visual embeddings with positional embedding fusion and a convolutional decoder, we aim to enable more accurate counting of complex, multi-part objects without the need for exemplars or textual guidance.

\section{Methodology}

In this paper, we address the challenging problem of visual object counting without reference images or arbitrary exemplars provided by the users. Our objective is to train a generalized visual object counter capable of performing effectively on a test set without any exemplars. The training dataset is defined as follows:
\begin{equation}
    D_{\text{train}} = \{(X_1,y_1), \ldots, (X_N,y_N)\}
\end{equation}
where $X_i \in R^{H \times W \times 3}$ represents the input image, and $y_i \in R^{H \times W \times 1}$ is a binary spatial density map. This map features 1's at the central location of the objects, denoting their presence, and 0's elsewhere where objects are absent. The total object count can be ascertained by performing a spatial summation over the density map. We aim to develop a visual object counter with a generalized approach, designed to work on a test set, even when provided with no exemplars, denoted as follows:
\begin{equation}
     D_{\text{test}} = \{X_{N+1}, \ldots, X_M\}.
\end{equation}
We employ a pretrained DINOv2 \cite{dinov2} vision foundation model as the visual encoder to extract self-supervised representations with strong spatial coherence. To preserve the spatial relationships inherent within the input images, we incorporate positional embeddings into the visual feature representations. The feature vectors, which now embody both the distilled knowledge and spatial context, are subsequently processed through a ConvNet-based decoder. This transformation yields a density map that is subsequently refined by a linear regression layer, resulting in a single-channel heatmap.

\subsection{Dataset \& Metrics}
The FSC-147 \cite{ranjan2021learning} dataset is widely used in class-agnostic counting, and contains 6135 images across 147 object classes.
The number of objects per image ranges from 7 to 3731, and averages at 56.
FSC-147 provides a corresponding binary density map $y_i$ for each training image $X_i$ in $D_{train}$.

We use the standard Mean Absolute Error (MAE) and Root Mean Squared Error (RMSE) to evaluate the performance of the model.
\begin{equation}
    MAE = \frac{1}{N} \sum_{i=1}^{N} |C_i - C_i^{GT}|,
\end{equation}
\begin{equation}
    RMSE = \sqrt{\frac{1}{N} \sum_{i=1}^{N} (C_i - C_i^{GT})^2}
\end{equation}
Here, $N$ refers to the number of images in the test set while $C_i$ and $C_i^{GT}$ refer to the predicted count and ground truth of the $i$-th image.
The count $C_i$ is obtained from summing the values across the density map $y_i$.

\begin{figure*}[!t]
\centering

\subfloat[GT: 21,\; CounTX: 15,\; Ours: 21]{
\begin{minipage}{0.48\textwidth}\centering
\includegraphics[width=0.31\textwidth,height=2cm]{figures/glasses_vs_countx/full_881.png}\hfill
\includegraphics[width=0.31\textwidth,height=2cm]{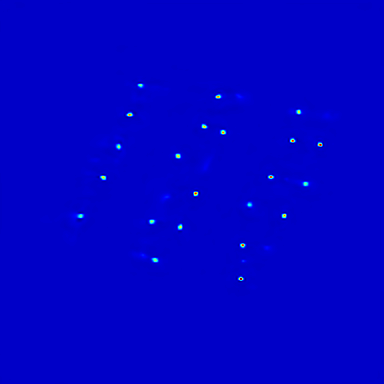}\hfill
\includegraphics[width=0.31\textwidth,height=2cm]{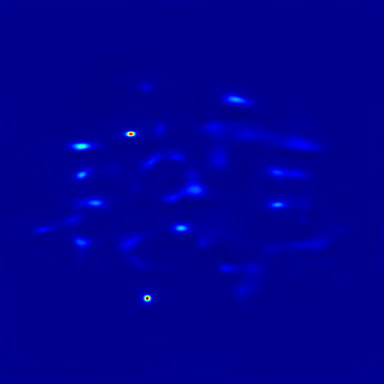}\\[-2pt]
\footnotesize Input \hspace{2.2cm} CounTX \hspace{2.2cm} Ours
\end{minipage}
}
\hfill
\subfloat[GT: 118,\; CounTX: 82,\; Ours: 118]{
\begin{minipage}{0.48\textwidth}\centering
\includegraphics[width=0.31\textwidth,height=2cm]{figures/perfect_count/full_866.png}\hfill
\includegraphics[width=0.31\textwidth,height=2cm]{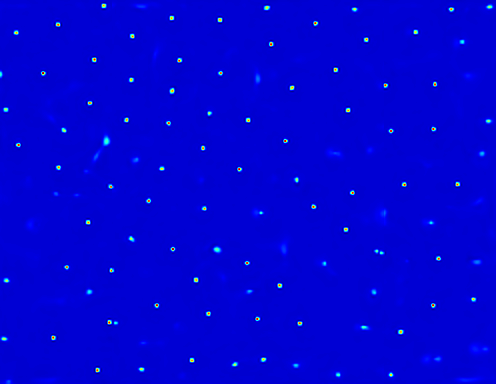}\hfill
\includegraphics[width=0.31\textwidth,height=2cm]{figures/perfect_count/density_map_866.png}\\[-2pt]
\footnotesize Input \hspace{2.2cm} CounTX \hspace{2.2cm} Ours
\end{minipage}
}

\vspace{6pt}

\subfloat[GT: 96,\; CounTX: 185,\; Ours: 122]{
\begin{minipage}{0.48\textwidth}\centering
\includegraphics[width=0.31\textwidth,height=2cm]{figures/glasses_vs_countx/full_1127.png}\hfill
\includegraphics[width=0.31\textwidth,height=2cm]{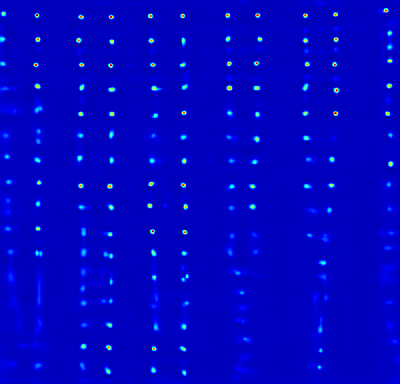}\hfill
\includegraphics[width=0.31\textwidth,height=2cm]{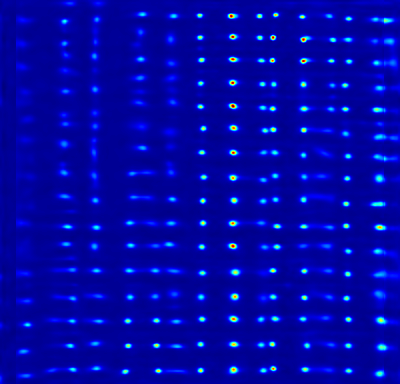}\\[-2pt]
\footnotesize Input \hspace{2.2cm} CounTX \hspace{2.2cm} Ours
\end{minipage}
}
\hfill
\subfloat[GT: 48,\; CounTX: 50,\; Ours: 48]{
\begin{minipage}{0.48\textwidth}\centering
\includegraphics[width=0.31\textwidth,height=2cm]{figures/perfect_count/full_928.png}\hfill
\includegraphics[width=0.31\textwidth,height=2cm]{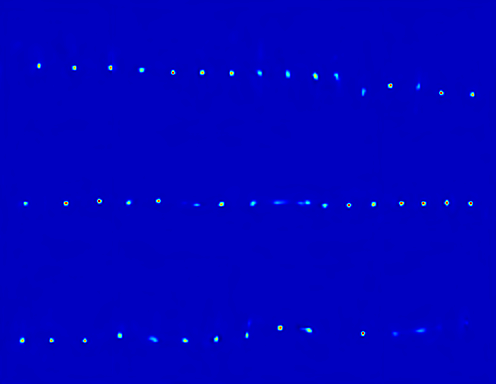}\hfill
\includegraphics[width=0.31\textwidth,height=2cm]{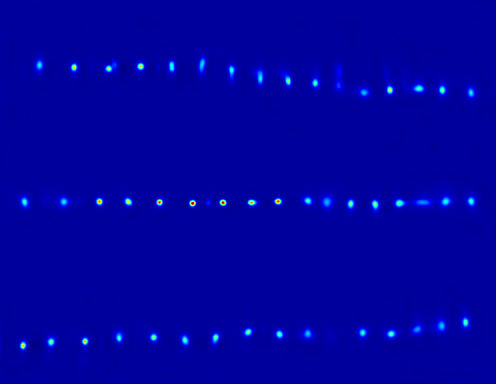}\\[-2pt]
\footnotesize Input \hspace{2.2cm} CounTX \hspace{2.2cm} Ours
\end{minipage}
}

\vspace{6pt}

\subfloat[GT: 79,\; CounTX: 143,\; Ours: 83]{
\begin{minipage}{0.48\textwidth}\centering
\includegraphics[width=0.31\textwidth,height=2cm]{figures/glasses_vs_countx/full_184.png}\hfill
\includegraphics[width=0.31\textwidth,height=2cm]{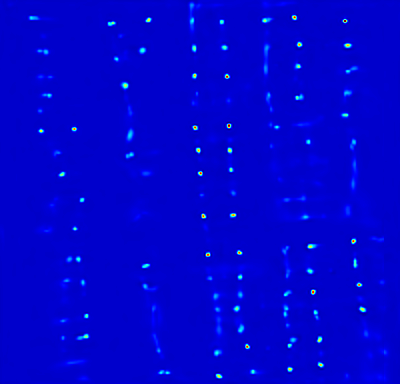}\hfill
\includegraphics[width=0.31\textwidth,height=2cm]{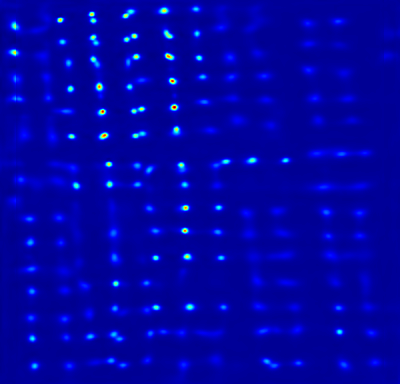}\\[-2pt]
\footnotesize Input \hspace{2.2cm} CounTX \hspace{2.2cm} Ours
\end{minipage}
}
\hfill
\subfloat[GT: 17,\; CounTX: 15,\; Ours: 17]{
\begin{minipage}{0.48\textwidth}\centering
\includegraphics[width=0.31\textwidth,height=2cm]{figures/perfect_count/full_111.png}\hfill
\includegraphics[width=0.31\textwidth,height=2cm]{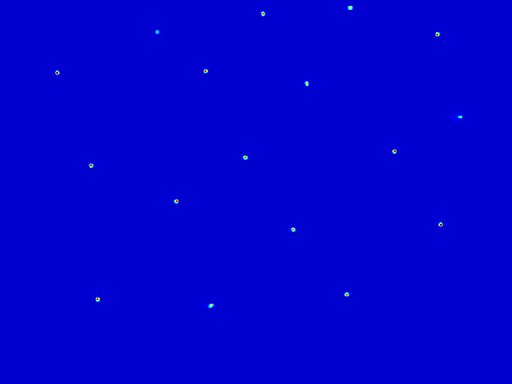}\hfill
\includegraphics[width=0.31\textwidth,height=2cm]{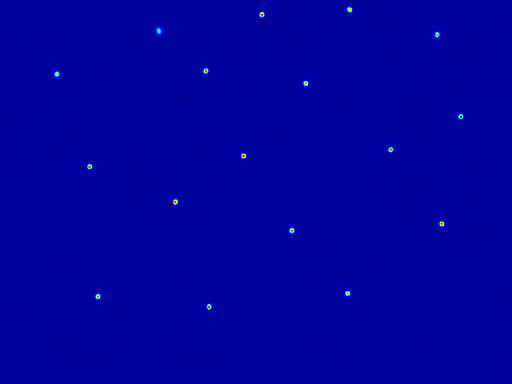}\\[-2pt]
\footnotesize Input \hspace{2.2cm} CounTX \hspace{2.2cm} Ours
\end{minipage}
}
\caption{Qualitative results of our model on FSC-147. Each sub-figure consists of the original image (left), CounTX's density map (middle), and our model's density map (right). The predicted counts from both models and ground truth (GT) counts have also been provided.}
\label{fig:qualitative}
\end{figure*}

\begin{table}[!t]
\centering
\caption{Comparison on FSC-147 under the standard split. 
MAE and RMSE are reported for validation and test sets. 
Our method is exemplar-free at inference, CounTX* uses text prompts.}
\label{tab:comparison}
\begin{tabular}{lcccc}
\hline
\textbf{Method} & \multicolumn{2}{c}{\textbf{Validation Set}} & \multicolumn{2}{c}{\textbf{Test Set}} \\
\cline{2-5}
 & \textbf{MAE} & \textbf{RMSE} & \textbf{MAE} & \textbf{RMSE} \\
\hline
GMN \cite{lu2019class} & 39.02 & 106.06 & 37.86 & 141.39 \\
FamNet \cite{ranjan2021learning} & 32.15 & 98.75 & 32.37 & 131.46 \\
CounTX* \cite{countx} & 17.70 & 63.61 & 15.73 & 106.88 \\
CounTR \cite{countr} & 18.07 & 71.84 & 14.71 & 106.87 \\
RCC \cite{hobley2022learning} & 17.49 & 58.81 & 17.12 & 104.53 \\
\textbf{Ours} & \textbf{19.11} & \textbf{65.58} & \textbf{19.06} & \textbf{118.45} \\
\hline
\end{tabular}
\end{table}

\subsection{Architecture}
In this section, we outline the architecture of our proposed model, illustrated in Figure~\ref{fig:architecture}. We use the DINOv2 visual encoder, denoted as $\mathcal{E}_{\mathtt{Encoder}}$, to convert the input image $\mathcal{X}_i$ into a feature map $\mathcal{F}_{\mathtt{DINO}}$. Positional embeddings, denoted as $\mathtt{E}$, are added to this feature map to obtain a spatially grounded representation $\mathcal{F}_{\mathtt{E}}$. A ConvNet-based decoder $\mathcal{D}_{\mathtt{Decoder}}$, consisting of four upsampling stages, then decodes $\mathcal{F}_{\mathtt{E}}$ into a one-channel density map $y_i$ at the original image resolution, representing the object density distribution. The overall computation is expressed in Equation~\ref{overall}.

\begin{equation}
    y_i = \mathcal{D}_{\mathtt{Decoder}}\!\left(\mathtt{E} + \mathcal{E}_{\mathtt{Encoder}}(\mathcal{X}_i)\right)
    \label{overall}
\end{equation}

\begin{figure*}[!t]
\centering
\subfloat[Input image]{
    \includegraphics[width=0.30\textwidth]{figures/countx_new/1127_original.png}
}\hfill
\subfloat[CounTX density map]{
    \includegraphics[width=0.30\textwidth]{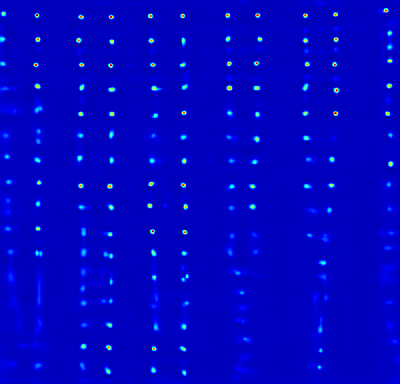}
}\hfill
\subfloat[Our model]{
    \includegraphics[width=0.30\textwidth]{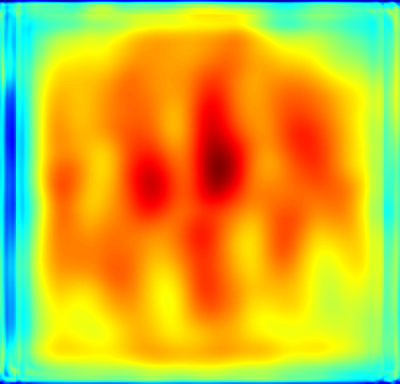}
}
\caption{Comparison on the “glasses” example from FSC-147. 
Ground truth (GT) count = 96; CounTX prediction = 185; our model’s prediction = 98. 
Left to right: input image, CounTX density map, and our model’s density map.}
\label{fig:glass_comparison}
\end{figure*}

\subsubsection{Visual Encoder} \label{encoder}
We adopt the pretrained DINOv2 \cite{dinov2} vision transformer as the visual backbone. DINOv2 is trained in a self-supervised manner using knowledge distillation\cite{hinton2015distilling,dino}, where a student network learns to align its representations with those of a teacher network across multiple image views. This training paradigm produces representations with strong semantic and spatial coherence.

In our framework, the encoder is used as a feature extractor without modifying its original training objective. For an input image $\mathcal{X}_i$, the encoder generates a token feature representation
\begin{equation}
    \mathcal{F}_{\mathtt{DINO}} = \mathcal{E}_{\mathtt{Encoder}}(\mathcal{X}_i) \in \mathbb{R}^{M \times D},
    \label{dinoencoder}
\end{equation}
where $M$ denotes the number of tokens and $D$ the embedding dimension.

\subsubsection{Positional Embedding}
To preserve spatial relationships between tokens, we add positional embeddings $\mathtt{E}$ to the visual features:
\begin{equation}
    \mathcal{F}_{\mathtt{E}} = \mathtt{E} + \mathcal{F}_{\mathtt{DINO}}.
\end{equation}
This fusion provides explicit spatial grounding before density decoding.

\subsubsection{Decoder}
The spatial features $\mathcal{F}_{\mathtt{E}}$ are reshaped into 2D feature maps and progressively upsampled using a ConvNet-based decoder $\mathcal{D}_{\mathtt{Decoder}}$:
\begin{equation}
    y_i = \mathcal{D}_{\mathtt{Decoder}}(\mathcal{F}_{\mathtt{E}}) \in \mathbb{R}^{H \times W \times 1}.
\end{equation}
The decoder consists of four upsampling stages followed by a linear layer producing the final density map.




\subsection{Implementation Details}
The input images are resized to a dimension of 256 x 256 pixels. Prior to model input, the images are transformed using center-cropping (224 x 224 pixels), horizontal and vertical flips, random rotations within 30 degrees, and finally, normalization. We employ the AdamW optimization method with a batch size set of 8 and a learning rate of 6.25e-6. We train our model with the RTX 3070 Ti Mobile GPU. After training for 200 epochs, the model exhibiting the lowest MAE on the validation set is chosen.

During inference time, we follow the same procedure as in \cite{countr}. Specifically, a square sliding window is adopted to scan over the image, using a stride of 128 pixels. Similar to the training phase, each image is resized to 256 x 256 pixels and subsequently normalized prior to being input into the model. For handling overlapping regions in the density map, we apply the approach described in \cite{countr}.

\section{Results and Discussion}

\subsection{Quantitative Results}

The evaluation results for our method on FSC-147 \cite{ranjan2021learning} are compared against other existing methods in Table \ref{tab:comparison}.
While not outperforming methods like CounTR \cite{countr} and RCC \cite{hobley2022learning}, our model performs reasonably well across both validation and test sets.

Although our overall MAE and RMSE are higher than CounTR and RCC under the standard FSC-147 protocol, our objective is not to optimize purely for aggregate error but to investigate structural robustness in exemplar-free counting. As shown in the qualitative results, our approach reduces part-level overcounting in composite objects (e.g., glasses), highlighting a complementary strength not fully captured by global MAE/RMSE metrics.

The most notable results are observed in the qualitative results in the next Section \ref{qualitative},
where our model demonstrates the ability to count more complex objects that CounTR fails to.

All quantitative results in Table~\ref{tab:comparison} follow the official FSC-147 split and include \emph{all} images. The exclusion of four extreme dense scenes is reported solely as a diagnostic sensitivity analysis (Table~\ref{tab:limitations}) to illustrate metric sensitivity, not as a modification of the benchmark protocol.

\begin{figure}[!t]
\centering
\subfloat[Input Lego Image]{
    \includegraphics[width=0.45\columnwidth]{figures/limitations/1102_original.png}
}\hfill
\subfloat[Our model’s density map]{
    \includegraphics[width=0.45\columnwidth]{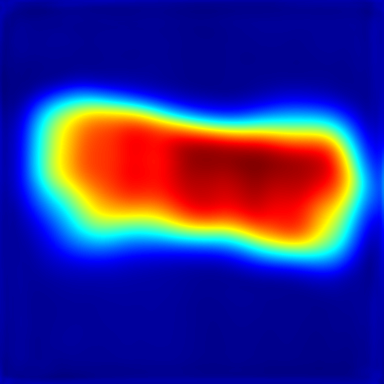}
}
\caption{Failure case on densely packed objects. 
Ground truth (GT) count = 512; our model’s prediction = 400. 
Left: input image. Right: predicted density map, illustrating reduced accuracy when objects lack clear boundaries.}
\label{fig:dense_lego_failure}
\end{figure}

\subsection{Qualitative Results} \label{qualitative}

Representative qualitative comparisons are shown in Figure~\ref{fig:qualitative}, where predictions from our model and CounTX \cite{countx} are visualized alongside ground truth counts. Notably, the density maps generated by our approach exhibit smoother spatial distributions, whereas CounTX produces activations that more closely resemble point-level responses.

The DINOv2 \cite{dinov2} backbone employed in our framework appears to encode structural regularities that influence counting behavior. In the ``glasses'' example shown in Figure~\ref{fig:glass_comparison}, competing methods such as CounTX \cite{countx} and CounTR \cite{countr} produce density responses concentrated around individual sub-components (e.g., lenses), leading to over-counting. In contrast, our model generates activations that are spatially aligned with whole-object instances, yielding predictions closer to the ground truth without requiring test-time augmentation.

As discussed in \cite{countr}, augmentation strategies can help mitigate certain structural ambiguities. Our observations suggest that representation-level differences—specifically the spatial coherence of DINOv2 features—may contribute to improved structural grouping in composite-object scenarios. These qualitative findings complement the quantitative analysis by illustrating behavioral differences not fully captured by aggregate MAE/RMSE metrics.

\subsection{Diagnostic Sensitivity Analysis on Extreme Dense Scenes}

To better understand the impact of extreme high-density cases, we conduct a focused diagnostic analysis on four images characterized by exceptionally large object counts and annotation ambiguity. These scenes represent rare but disproportionately influential samples within the dataset.

Importantly, these images are \emph{not} excluded from the official benchmark evaluation. All primary results reported in Table~\ref{tab:comparison} strictly follow the standard FSC-147 split and include all validation and test images. Table~\ref{tab:limitations} summarizes the official benchmark results alongside the diagnostic variant excluding the four extreme cases. The exclusion is applied only for analytical purposes to illustrate how a small number of extreme-density cases can substantially inflate squared-error metrics, particularly RMSE. This analysis is intended to clarify metric sensitivity rather than modify the evaluation protocol.

\begin{table}[!t]
\centering
\caption{Diagnostic sensitivity analysis on extreme dense-scene cases in FSC-147.}
\label{tab:limitations}
\setlength{\tabcolsep}{4pt}
\begin{tabular}{lcc|cc}
\hline
\textbf{Config.} & \multicolumn{2}{c|}{\textbf{Val}} & \multicolumn{2}{c}{\textbf{Test}} \\
 & MAE & RMSE & MAE & RMSE \\
\hline
Official & 19.11 & 65.58 & 19.06 & 118.45 \\
Diagnostic & 15.64 & 48.12 & 13.14 & 33.05 \\
\hline
\end{tabular}
\end{table}

\subsection{Model Limitations and Future Directions}

Our model exhibits clear weaknesses in extremely dense scenes with many small, visually similar objects and weak boundaries (e.g., tightly packed Lego pieces in Figure~\ref{fig:dense_lego_failure}), where overlapping density activations lead to systematic underestimation.

These errors are influenced not only by representation quality but also by input resolution (224$\times$224) and decoder capacity, which may suppress fine-grained boundary details. In addition, occasional annotation granularity ambiguity (e.g., stud-level versus piece-level counting) introduces evaluation sensitivity and disproportionately affects RMSE.

Because we intentionally retain the density-regression objective and decoder structure of prior work to isolate representation-level effects, architectural modifications were not explored. Future work will investigate higher-resolution inputs, multi-scale aggregation, and broader cross-dataset evaluation to better address dense and structurally ambiguous scenarios.

\section{Conclusion}

We presented CountFormer, a controlled integration of a self-supervised vision foundation model (DINOv2) into an exemplar-free density-regression framework for class-agnostic object counting. By combining foundation-level representations with explicit positional embedding fusion and a lightweight decoder, we examined how representation quality influences structural consistency without modifying the underlying counting objective or benchmark protocol. Experiments on FSC-147 demonstrate competitive performance among zero-shot and exemplar-free methods under the official evaluation split. While aggregate MAE and RMSE remain broadly aligned with prior approaches, qualitative results indicate improved structural grouping in certain composite-object scenarios, suggesting that foundation-based representations can influence part–whole coherence in counting tasks. Overall, this study highlights the importance of representation design in exemplar-free counting and provides a controlled baseline for further exploration of structure-aware visual counting systems.

{
\bibliographystyle{IEEEtran}
\bibliography{fsl}
}
\end{document}